\pdfoutput=1

\documentclass[11pt]{article}

\usepackage{acl}

\usepackage{times}
\usepackage{latexsym}

\usepackage[T1]{fontenc}

\usepackage[utf8]{inputenc}

\usepackage{soul}
\usepackage{url}
\usepackage{ulem}
\usepackage{booktabs}
\usepackage{amsmath}
\usepackage{amsthm} 
\usepackage{amsfonts}  
\usepackage{amssymb}
\usepackage{multicol}
\usepackage{multirow}
\usepackage{graphicx}
\usepackage{booktabs}
\usepackage{colortbl}
\usepackage{stfloats}
\usepackage{comment}
\usepackage{makecell}
\usepackage[noend]{algpseudocode}
\usepackage{algorithmicx,algorithm}
\usepackage{pifont}
\usepackage{color}
\usepackage{microtype}
\usepackage{cleveref}

\title{DARER: Dual-task Temporal Relational Recurrent Reasoning Network \\for Joint Dialog Sentiment Classification and Act Recognition
}

\author{Bowen Xing$^1$ and Ivor W. Tsang$^{2,1}$ \\ 
$^1$Australian Artificial Intelligence Institute, University of Technology Sydney, Australia \\
$^2$Centre for Frontier Artificial Intelligence Research, A*STAR, Singapore\\ 
\texttt{\normalsize{bwxing714@gmail.com, ivor\_tsang@ihpc.a-star.edu.sg}}\\
\texttt{}}
\begin{document}
\maketitle
\begin{abstract}
The task of joint dialog sentiment classification (DSC) and act recognition (DAR) aims to simultaneously predict the sentiment label and act label for each utterance in a dialog.
In this paper, we put forward a new framework 
 which models the explicit dependencies via integrating \textit{prediction-level interactions} other than semantics-level interactions, more consistent with human intuition.
Besides, we propose a speaker-aware temporal graph (SATG) and a dual-task relational temporal graph (DRTG) to introduce \textit{temporal relations} into dialog understanding and dual-task reasoning. 
To implement our framework, we propose a novel model
dubbed DARER, which first generates the context-, speaker- and temporal-sensitive utterance representations via modeling SATG, then conducts recurrent dual-task relational reasoning on DRTG, in which process the estimated label distributions act as key clues in prediction-level interactions.
Experiment results show that DARER outperforms existing models by large margins while requiring much less computation resource and costing less training time.
Remarkably, on DSC task in Mastodon, DARER gains a relative improvement of about 25\% over previous best model in terms of F1, with less than 50\% parameters and about only 60\% required GPU memory.

\end{abstract} 
\section{Introduction}\label{sec:introduction}
Dialog sentiment classification (DSC) and dialog act recognition (DAR) are two challenging tasks in dialog systems \cite{empiricalstudyondialog}.
DSC aims to predict the sentiment label of each utterance in a dialog, while DAR aims to predict the act label.
Recently, researchers have discovered that these two tasks are correlative and they can assist each other \cite{mastodon,kimkim}.

An example is shown in Table \ref{table: sample}.
To predict the sentiment of $u_b$, besides its \textit{semantics}, its Disagreement act \textit{label} and the Positive sentiment \textit{label} of its \textit{previous} utterance ($u_a$) can provide useful references, which contribute a lot when humans do this task.
This is because the Disagreement act label of $u_b$ denotes it has the opposite opinion with $u_a$, and thus $u_b$ tends to have a Negative sentiment label, the opposite one with $u_a$ (Positive).
Similarly, the opposite sentiment labels of $u_b$ and $u_a$ are helpful to infer the Disagreement act label of $u_b$.
In this paper, we term this process as dual-task reasoning, where there are three key factors: 
1) the semantics of $u_a$ and $u_b$; 2) the temporal relation between $u_a$ and $u_b$; 3) $u_a$'s and $u_b$'s labels for another task.

\begin{table}[t]
\centering
\fontsize{8}{11}\selectfont
\setlength{\tabcolsep}{0.4mm}{
\begin{tabular}{l|l|l}
\toprule
 Utterances& Act& Sentiment \\ \midrule
 \begin{tabular}[c]{@{}l@{}}$u_a$: I highly recommend it. Really awe-\\some progression and added difficulty \end{tabular} 
& Statement & Positive \\\hline
 $u_{b}$: I never have.
&Disagreement &Negative \\
\bottomrule
\end{tabular}}
\caption{A dialog snippet from the Mastodon dataset.}
\label{table: sample}
\end{table}

\begin{figure}[t]
 \centering
 \includegraphics[width = 0.38\textwidth]{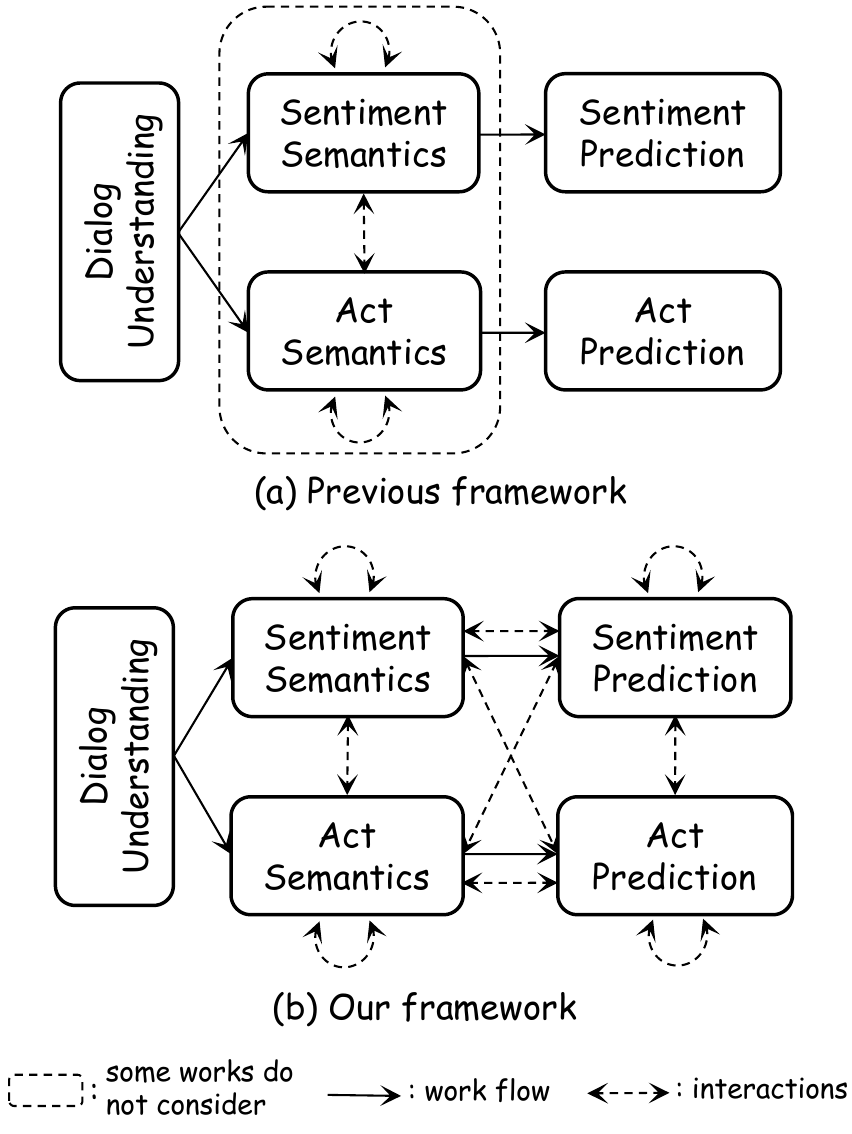}
 \caption{Illustration of previous framework and ours.}
 \label{fig: framework}
\end{figure}

In previous works, different models are proposed to model the correlations between the two tasks.
\cite{mastodon} propose a multi-task model in which the two tasks share a single encoder.
\cite{kimkim,dcrnet,bcdcn,cogat} try to model the semantics-level interactions of the two tasks.
The framework of previous models is shown in Fig. \ref{fig: framework} (a).
For dialog understanding, 
Co-GAT \cite{cogat} applies graph attention network (GAT) \cite{gat}
over an undirected disconnected graph which consists of isolated speaker-specific full-connected subgraphs.
Therefore, it suffers from the issue that the inter-speaker interactions cannot be modeled, and the temporal relations between utterances are omitted.
For dual-task reasoning, on the one hand,
previous works only consider the parameter sharing and semantics-level interactions, while the label information is not integrated into the dual-task interactions.
Consequently, 
the explicit dependencies between the two tasks cannot be captured
and previous dual-task reasoning processes are inconsistent with human intuition, which leverages the label information as crucial clues.
On the other hand, previous works do not consider the temporal relations between utterances in dual-task reasoning, while in which they play a key role.

In this paper, we try to address the above issues by introducing temporal relations and leveraging label information. 
To introduce temporal relations, we design a \textbf{s}peaker-\textbf{a}ware \textbf{t}emporal \textbf{g}raph (SATG) for dialog understanding, and a \textbf{d}ual-task \textbf{r}easoning \textbf{t}emporal \textbf{g}raph (DRTG) for dual-task relational reasoning.
Intuitively, different speakers' semantic states will change as the dialog goes, and these semantic state transitions trigger different sentiments and acts.
SATG is designed to model the speaker-aware semantic states transitions, which provide essential indicative semantics for both tasks.
Since the temporal relation is a key factor in dual-task reasoning,
DRTG is designed to integrate inner- and inter-task temporal relations, making the dual-task reasoning process more rational and effective.

To leverage label information, we propose a new framework, as shown in Fig. \ref{fig: framework} (b).
Except for semantics-level interactions, it integrates several kinds of prediction-level interactions.
First, self-interactions of sentiment predictions and act predictions.
In both tasks, there are prediction-level correlations among the utterances in a dialog.
In the DSC task, the sentiment state of each speaker tends to be stable until the utterances from others trigger the changes \cite{dialoggcn,ercseqtag}.
In the DAR task, there are different patterns (e.g., Questions-Inform and Directives-Commissives) reflecting the interactions between act labels \cite{dailydialog}. 
Second, interactions between the predictions and semantics.
Intuitively, the predictions can offer feedback to semantics, which can rethink then reversely help revise the predictions.
Third, prediction-prediction interactions between DSC and DAR, which model the explicit dependencies.
However, since our objective is to predict the labels of both tasks, there is no ground-truth label available for prediction-level interactions.
To this end, we design a recurrent dual-task reasoning mechanism that leverages the label distributions estimated in the previous step as prediction clues of the current step for producing new predictions.
In this way, the label distributions of both tasks are gradually improved along the step.

To implement our framework, we propose a novel \textbf{D}ual-t\textbf{A}sk temporal \textbf{R}elational r\textbf{E}current \textbf{R}easoning Network (DARER), which includes three modules.
The Dialog Understanding module conducts relation-specific graph transformations (RSGT) over SATG to generate context-, speaker- and temporal-sensitive utterance representations.
The Initial Estimation module outputs the initial label information fed to the Recurrent Dual-task Reasoning module, in which RSGT operates on DRTG to conduct dual-task relational reasoning.
Moreover, we design logic-heuristic training objectives to force DSC and DAR to prompt each other in the recurrent dual-task reasoning process gradually.
Experiments on public datasets show that DARER significantly outperforms existing models.
And further improvements can be obtained by utilizing pre-trained language models as the utterance encoder.
Besides, compared with the previous best model, DARER reduces the number of parameters, required GPU memory, and training time.

The source code of DARER is publicly available at \url{https://github.com/XingBowen714/DARER}.


\section{Methodology}
Given a dialog consisting of N utterances: $\mathcal D\!=\!(u_1, u_2, ..., u_N$), our objective is to predict both the dialog sentiment labels $Y^S \!=\! {y^s_1, ..., y^s_N}$ and the dialog act labels $Y^A \!= \!{y^a_1, ..., y^a_N}$ in a single run.
Before delving into the details of DARER, we start with our designed SATG and DRTG.
\subsection{Speaker-aware Temporal Graph}
We design a SATG to model the information aggregation between utterances in a dialog.
Formally, SATG is a complete directed graph denoted as $\mathcal{G}=(\mathcal{V,E,R})$.
In this paper, the nodes in $\mathcal{G}$ are the utterances in the dialog, i.e., $\left|\mathcal{V}\right|=N,\mathcal{V}=(u_1, ..., u_N)$, and the edge $(i,j,r_{ij})\in \mathcal{E}$ denotes the information aggregation from $u_i$ to $u_j$ under the relation $r_{ij}\in \mathcal{R}$.
Table \ref{table: sagraph} lists the definitions of all relation types in $\mathcal{R}$.
In particular, there are three kinds of information conveyed by $r_{ij}$: the speaker of $u_i$, the speaker of $u_j$, and the relative position of $u_i$ and $u_j$.
Naturally, the utterances in a dialog are chronologically ordered, so the relative position of two utterances denotes their temporal relation.
An example of SATG is shown in Fig. \ref{table: sagraph}.
Compared with previous dialog graph structure \cite{dcrnet,cogat}, our SATG has two main advancements.
First, as a complete directed graph,
SATG can model both the intra- and inter-speaker semantic interactions. 
Second, incorporating temporal information, SATG can model the transitions of speaker-aware semantic states as the dialog goes on, which benefit both tasks.
\begin{table}[t]
\centering
\fontsize{9}{11}\selectfont
\setlength{\tabcolsep}{1.5mm}{
\begin{tabular}{c|cccccccc}
\toprule
$r_{ij}$  &1&2&3&4&5&6&7&8   \\ \midrule
$I_s(i)$ &1&1&1&1&2&2&2&2 \\
$I_s(j)$ &1&1&2&2&1&1&2&2\\
$pos(i,j)$&$>$&$\leq$&$>$&$\leq$&$>$&$\leq$&$>$&$\leq$\\
\bottomrule
\end{tabular}}
\caption{All relation types in SATG (assume there are two speakers). $I_s(i)$ indicates the speaker node $i$ is from. $pos(i,j)$ indicates the relative position of node $i$ and $j$.}
\label{table: sagraph}
\end{table}
\begin{figure}[t]
 \centering
 \includegraphics[width = 0.4\textwidth]{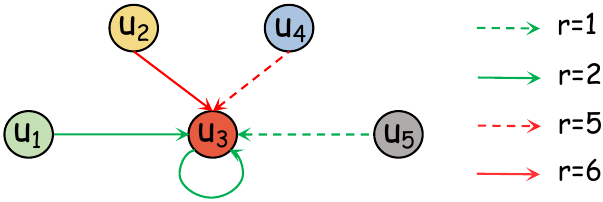}
 \caption{An example of SATG. $u_1, u_3$ and $u_5$ are from speaker 1 while $u_2$ and $u_4$ are from speaker 2.
 w.l.o.g, only the edges directed into $u_3$ node are illustrated.}
 \label{fig: sagraph}
\end{figure}

\subsection{Dual-task Reasoning Temporal Graph}
We design a DRTG to provide an advanced platform for dual-task relational reasoning.
It is also a complete directed graph that consists of $2N$ dual nodes: $N$ sentiment nodes and $N$ act nodes.
The definitions of all relation types in $\mathcal{R'}$ are listed in Table \ref{table: dtgraph}.
Intuitively, when predicting the label of a node, the information of its dual node plays a key role, so we emphasize the temporal relation of `$=$' rather than merge it with `$<$' like SATG. 
Specifically, the relation $r'_{ij}$ conveys three kinds of information: the task of $n_i$, the task of $n_j$ and the temporal relation between $n_i$ and $n_j$.
An example of DRTG is shown in Fig. \ref{table: dtgraph}.
Compared with previous dual-task graph structure \cite{dcrnet,cogat}, our DRTG has two major advancements.
First, the temporal relations in DRTG can make the DTR-RSGT capture the the temporal information, which are essential for dual-task reasoning, while this cannot be achieved by the co-attention \cite{dcrnet} or graph attention network \cite{cogat} operating on their non-temporal graphs.
Second, in DRTG , the information aggregated into a node is decomposed by different relations that correspond to individual contributions, rather than only depending on the semantic similarity measured by the attention mechanisms.
\begin{table}[t]
\centering
\fontsize{9}{11}\selectfont
\setlength{\tabcolsep}{1mm}{
\begin{tabular}{c|cccccccccccc}
\toprule
$r'_{ij}$ &1& 2& 3& 4& 5& 6& 7& 8& 9&10&11&12     \\ \midrule
$I_t(i)$          & S&S&S&S&S&S&A&A&A&A&A&A\\
$I_t(j)$          &S&S&S&A&A&A&S&S&S&A&A&A\\
$pos(i, j)$&$<$&$=$ &$>$&$<$&$=$&$>$&$<$&$=$&$>$&$<$&$=$ &$>$\\
\bottomrule
\end{tabular}}
\caption{All relation types in DRTG. $I_t(i)$ indicates that node $i$ is a sentiment (S) node or act (A) node.}
\label{table: dtgraph}
\end{table}
\begin{figure}[t]
 \centering
 \includegraphics[width = 0.45\textwidth]{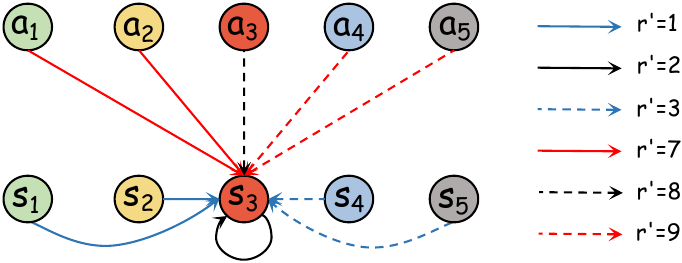}
 \caption{An example of DRTG. $s_i$ and $a_i$ respectively denote the node of DAC task and DAR task. w.l.o.g, only the edges directed into $s_3$ are illustrated.}
 \label{fig: dtgraph}
\end{figure}
\subsection{DARER}
\begin{figure*}[t]
 \centering
 \includegraphics[width = \textwidth]{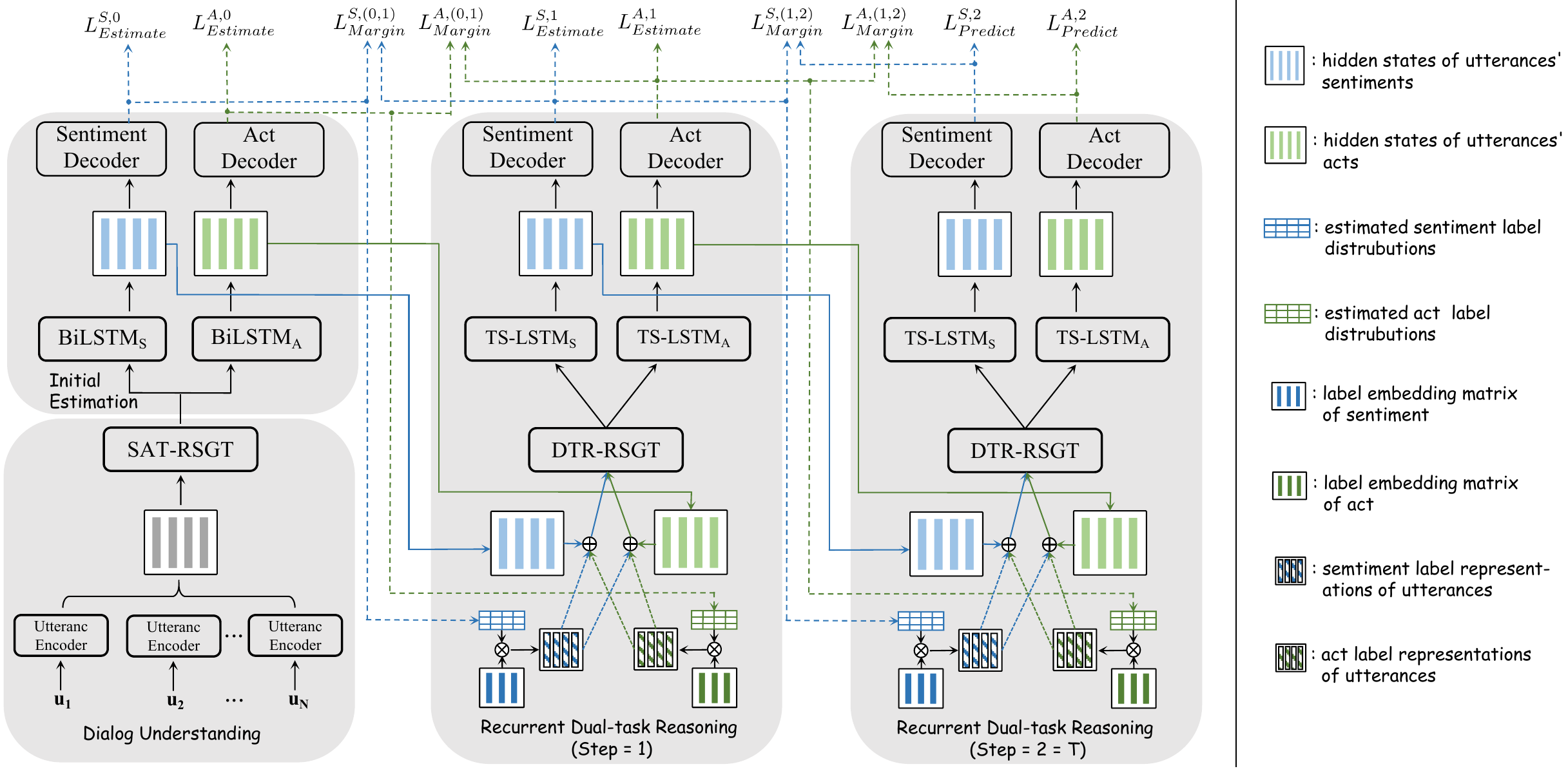}
 \caption{The network architecture of our proposed DARER model. Without loss of generality, the step number $T$ in this illustration is set 2.}
 \label{fig: model}
\end{figure*}
The network architecture of our proposed DARER is shown in Fig. \ref{fig: model}.
It consists of three modules, and we introduce their details next.
\subsubsection{Dialog Understanding}
\paragraph{Utterance Encoding}
In previous works, BiLSTM \cite{LSTM} is widely adopted as the utterance encoder to generate the initial utterance representation: $H=(h_0,...,h_N)$.
In this paper, besides BiLSTM, we also study the effect of different pre-trained language model (PTLM) encoders in Sec. \ref{sec: ptlm}.\\
\textbf{BiLSTM:} We apply the BiLSTM over the word embeddings of $u_t$ to capture the inner-sentence dependencies and temporal relationships among the words, producing a series of hidden states $H_{u,i} = (h_{u,i}^0, ..., h_{u,i}^{l_i})$, where $l_i$ is the length of $u_i$.
Then we feed $H_{u,i}$ into a max-pooling layer to get the representation for each $u_i$.\\
\textbf{PTLM:} We separately feed each utterance into the PTLM encoder and take the output hidden state of the \texttt{[CLS]} token as the utterance representation.

\paragraph{Speaker-aware Temporal RSGT}
To capture the inter- and intra-speaker semantic interactions and the speaker-aware temporal dependencies between utterances, we conduct Speaker-aware Temporal relation-specific graph transformations (SAT-RSGT) inspired from \cite{rgcn} over SATG. 
The information aggregation of SAT-RSGT can be formulated as: 
\begin{equation}
\hat{h}_{i}=W_{1} h_{i}^0 + \sum_{r \in \mathcal{R}} \sum_{j \in \mathcal{N}_{i}^{r}} \frac{1}{\left|{N}_{i}^{r} \right|} W^r_{1} h_{j}^0
\end{equation}
where $W_{1}$ is self-transformation matrix and $W_1^r$ is relation-specific matrix.
Now we obtain the context-, speaker- and temporal-sensitive utterance representations: $\hat{H}=(\hat{h_0},...,\hat{h_N})$.
\subsubsection{Initial Estimation}
To obtain task-specific utterances representations, we separately apply two BiLSTMs over $\hat{H}$ to obtain the utterance hidden states for sentiments and acts respectively: $H_s^0=\text{BiLSTM}_\text{S}(\hat{H})$, $H_a^0=\text{BiLSTM}_\text{A}(\hat{H})$, where $H_s^0=\{h_{s,i}^0\}_{i=1}^{N}$ and $H_a^0=\{h_{a,i}^0\}_{i=1}^{N}.$
Then $H_s^0$ and $H_a^0$ are separately fed into Sentiment Decoder and Act Decoder to produce the initial estimated label distributions:
 \begin{equation}
 \begin{split}
  P^0_{S}&=\{ P^0_{S,i}\}_{i=1}^N, \ P^0_{A}=\{ P^0_{A,i}\}_{i=1}^N\\
 P^0_{S,i}
 &= softmax(W_d^s h^0_{a,i} + b_d^s)\\
 &= \left[p^0_{s,i}[0], ..., p^0_{s,i}[k], ..., p^0_{s,i}(\left|\mathcal{C}_s\right|\!-\!1)\right]\\
 P^0_{A,i} 
 &= softmax(W_d^a h^0_{s,i} + b^a_d)\\
 &=  \left[p^0_{a,i}[0], ..., p^0_{a,i}[k], ..., p^0_{a,i}(\left|\mathcal{C}_a\right|\!-\!1)\right]
\end{split}
\end{equation}
where $W_d^*$ and $b_d^*$ are weight matrices and biases, $\mathcal{C}_s$ and $\mathcal{C}_a$ are sentiment class set and act class set.

\subsubsection{Recurrent Dual-task Reasoning}
At step $t$, the recurrent dual-task reasoning module takes two streams of inputs: 1) hidden states $H_s^{t-1}\in\mathbb{R}^{N\times d}$ and $H_a^{t-1}\in\mathbb{R}^{N\times d}$; 2) label distributions $P^{t-1}_{S}\in\mathbb{R}^{N \times \left|\mathcal{C}_s\right|}$ and $P^{t-1}_{A}\in\mathbb{R}^{N \times \left|\mathcal{C}_a\right|}$.

\paragraph{Projection of Label Distribution}
To achieve the prediction-level interactions, we should represent the label information in vector form to let it participate in calculations.
We use $P_S^{t-1}$ and $P_A^{t-1}$ to respectively multiply the sentiment label embedding matrix $M^e_s\in\!\mathbb{R}^{\left|\mathcal{C}_s\right|\times d}$ and the act label embedding matrix $M^e_a\in\! \mathbb{R}^{\left|\mathcal{C}_a\right|\times d}$, obtaining the sentiment label representations $E_S^{t}=\{e_{s,i}^t\}_{i=1}^N$ and act label representations $E_A^{t}=\{e_{a,i}^t\}_{i=1}^N$.
In particular, for each utterance, its sentiment label representation and act label representation are computed as:
 \begin{equation}
 \begin{split}
e_{s,i}^t =& \sum_{k=0}^{\left|\mathcal{C}_s\right|-1} p^{t-1}_{s,i}[k] \cdot v_s^k\\
 e_{a,i}^t =& \sum_{k'=0}^{\left|\mathcal{C}_a\right|-1} p^{t-1}_{a,i}[k'] \cdot v_a^{k'}
\end{split}
\end{equation}
where $v_s^k$ and $v_a^{k'}$ are the label embeddings of sentiment class $k$ and act class $k'$, respectively.

\paragraph{Dual-task Reasoning RSGT}
To achieve the self- and mutual-interactions between the semantics and predictions, for each node in DRTG, we superimpose its corresponding utterance's 
label embeddings of both tasks on its hidden state: 
 \begin{equation}
 \begin{split}
\hat{h}_{s,i}^t =& h_{s,i}^{t-1} + e_{s,i}^t + e_{a,i}^t \\
\hat{h}_{a,i}^t =& h_{a,i}^{t-1} + e_{s,i}^t + e_{a,i}^t
\end{split}
\end{equation}
Thus the representation of each node contains the task-specific semantic features and both tasks' label information, which are then incorporated into the relational reasoning process to achieve semantics-level and prediction-level interactions. 

The obtained $\mathbf{\hat{H}_s^t}$ 
and $\mathbf{\hat{H}_a^t}$ 
both have $N$ vectors, respectively corresponding to the $N$ sentiment nodes and $N$ act nodes on DRTG.
Then we feed them into the Dual-task Reasoning relation-specific graph transformations (DTR-RSGT) conducting on DRTG.
Specifically, the node updating process of DTR-RSGT can be formulated as:
\begin{equation}
\overline{h}_{i}^{t}=W_{2} \hat{h}_{i}^t + \sum_{r \in \mathcal{R'}} \sum_{j \in \mathcal{N}_{i}^{r'}} \frac{1}{\left|{N}_{i}^{r'} \right|} W^r_{2} \hat{h}_{j}^t
\end{equation}
where $W_{2}$ is self-transformation matrix and $W_2^r$ is relation-specific matrix.
Now we get $\overline{H}_s^t$ and $\overline{H}_a^t$.
\paragraph{Label Decoding}
For each task, we use a task-specific BiLSTM (TS-BiLSTM) to generate a new series of hidden states that are more task-specific:
 \begin{equation}
 \begin{split}
{H}_s^t&=\text{TS-BiLSTM}_\text{S}(\overline{H}_s^t) \\
{H}_a^t&= \text{TS-BiLSTM}_\text{A}(\overline{H}_a^t)
\end{split}
\end{equation}
Besides, as $\overline{H}_s^t$ and $\overline{H}_a^t$ both contains the label information of the two tasks, the two $\text{TS-BiLSTM}$s have another advantage of label-aware sequence reasoning, which has been proven can be achieved by LSTM \cite{lstmsoftmax}. 

Then ${H}_S^t$ and ${H}_A^t$ are separately fed to Sentiment Decoder and Act Decoder to produce $P^{t}_{S}$ and $P^{t}_{A}$.

\subsubsection{Logic-heuristic Training Objective}
Intuitively, there are two important logic rules in our DARER.
First, the produced label distributions should be good enough to provide useful label information for the next step.
Otherwise, noisy label information would be introduced, misleading the dual-task reasoning.
Second, both tasks are supposed to learn more and more beneficial knowledge from each other in the recurrent dual-task reasoning process. Scilicet the estimated label distributions should be gradually improved along steps.
In order to force DARER to obey these two rules, we propose a constraint loss $L_{Constraint}$ that includes two terms: $L_{Estimate}$ and $L_{Margin}$,  which correspond to the two rules, respectively.
\paragraph{Estimate Loss}
$L_{Estimate}$ is the cross-entropy loss forcing DARER to provide good enough label distributions for the next step.
At step $t$, for DSC task, $\mathcal{L}^{S,t}_{Estimate}$ is defined as:
\begin{equation}
\mathcal{L}^{S,t}_{Estimate}=\sum_{i=1}^{N}\sum_{k=0}^{\left|\mathcal{C}_s\right|-1}y_{s,i}^k \text{log}\left(p_{s,i}^{t}[k]\right) \label{eq: estloss}
\end{equation}

\paragraph{Margin Loss}
$L_{Margin}$ works on the label distributions of two adjacent steps, and it promotes the two tasks gradually learning beneficial knowledge from each other via forcing DARER to produce better predictions at step $t$ than step $t-1$.
For DSC task, $\mathcal{L}^{S,(t,t-1)}_{Margin}$ is a margin loss defined as:
\begin{equation}
\mathcal{L}^{S,(t,t-1)}_{Margin}\!=\!\sum_{i=1}^{N}\sum_{k=0}^{\left|\mathcal{C}_s\right|-1}\!\!y_{s,i}^k \ \text{max}(0,p_{s,i}^{t-1}[k]-p_{s,i}^{t}[k])\label{eq: marginloss}
\end{equation}
\paragraph{Constraint loss}
$L_{Constraint}$ is the weighted sum of $L_{Estimate}$ and  $L_{Margin}$, with a hyper-parameter $\gamma$ balancing the two kinds of punishments.
For DSC task, $\mathcal{L}^S_{Constraint}$ is defined as:
\begin{equation}
\mathcal{L}^S_{Constraint}=\sum_{t=0}^{T-1}\mathcal{L}^{S,t}_{Estimate} + \gamma * \sum_{t=1}^{T}\mathcal{L}^{S,(t,t-1)}_{margin}\label{eq: Constraintloss}
\end{equation}
\paragraph{Final Training Objective}
The total loss for DSC task ($\mathcal{L}^S$) is the sum of $\mathcal{L}^S_{Constraint}$ and $\mathcal{L}^S_{Prediction}$:
\begin{equation}
 \mathcal{L}^S =\mathcal{L}^S_{Prediction} + \mathcal{L}^S_{Constraint} \label{eq: sentiloss}
\end{equation}
where $\mathcal{L}^S_{Prediction}$ is the cross-entropy loss of the produced label distributions at the final step $T$:
\begin{equation}
\mathcal{L}^S_{Prediction}=\sum_{i=1}^{N}\sum_{k=0}^{\left|\mathcal{C}_s\right|-1}y_{s,i}\ \text{log}\left(p_{s,i}^{T}[k]\right) \label{eq: predloss}
\end{equation}

The total loss of DAR task ($\mathcal{L}^A$) can be derivated similarly like \cref{eq: estloss,eq: marginloss,eq: Constraintloss,eq: sentiloss,eq: predloss}.

The final training objective of DARER is the sum of the total losses of the two tasks:
\begin{equation}
\mathcal{L} = \mathcal{L}^S+ \mathcal{L}^A
\end{equation}
\section{Experiments}
\subsection{Datasets and Metrics}
\begin{figure}[t]
 \centering
 \includegraphics[width = 0.48\textwidth]{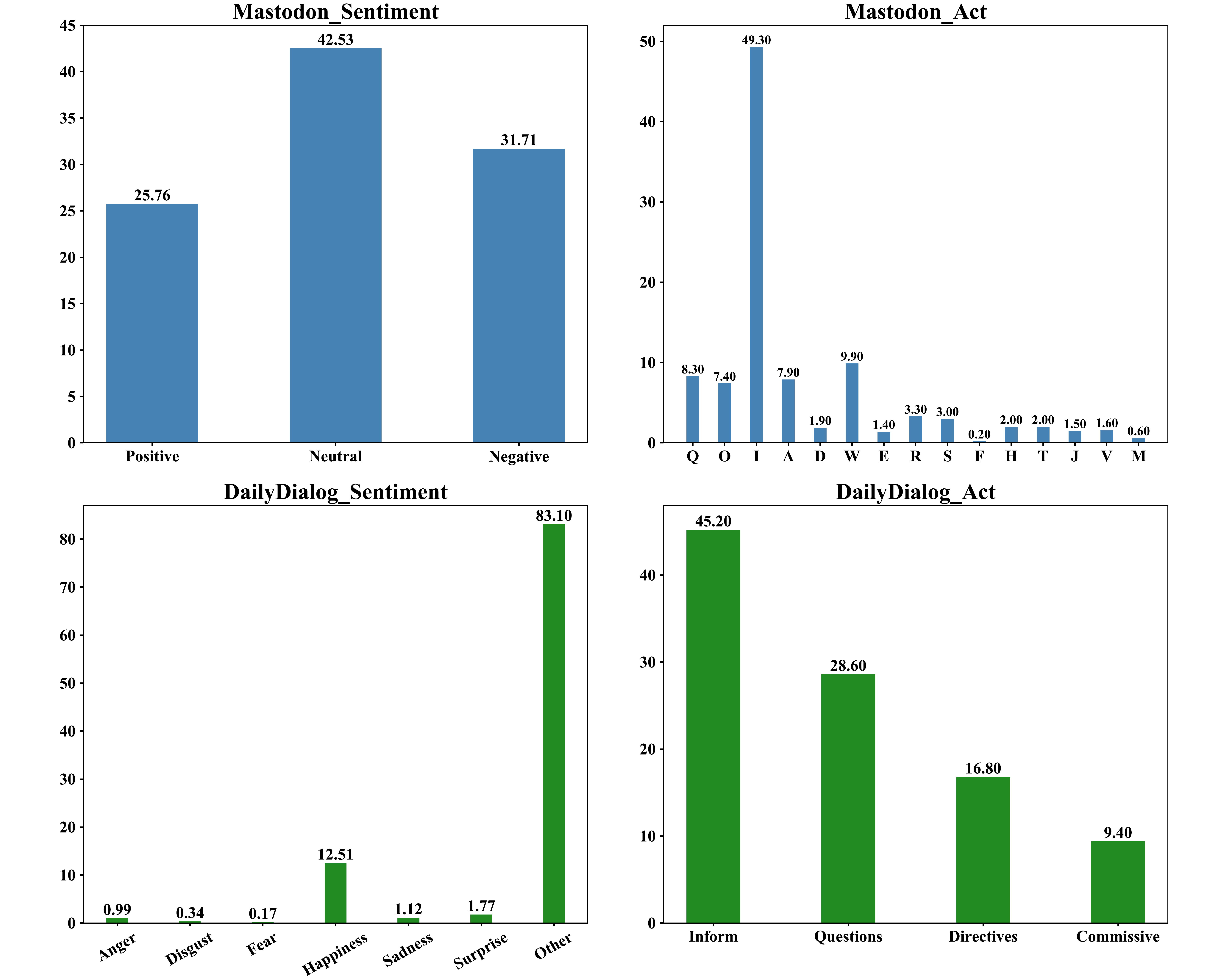}
 \caption{Illustration of class distributions.}
 \label{fig: label_distrib}
\end{figure}

\textbf{Dataset}. We conduct experiments on two publicly available dialogue datasets: Mastodon\footnote{https://github.com/cerisara/DialogSentimentMastodon} \cite{mastodon} and Dailydialog\footnote{http://yanran.li/dailydialog} \cite{dailydialog}.
The Mastodon dataset includes 269 dialogues for training and 266 dialogues for testing. 
And there are 3 sentiment classes and 15 act classes.
Since there is no official validation set, we follow the same partition as \citet{cogat}.
Finally, there are 243 dialogues for training, 26 dialogues for validating, and 266 dialogues for testing.
As for Dailydialog dataset, we adopt the official train/valid/test/ split from the original dataset \cite{dailydialog}: 11,118 dialogues for training, 1,000 for validating, and 1,000 for testing. And there are 7 sentiment classes and 4 act classes.
The class distributions of the two tasks on the two datasets are illustrated in Fig. \ref{fig: label_distrib}.\\
\textbf{Evaluation Metrics}. Following previous works \cite{mastodon,dcrnet,cogat}, on Dailydialog dataset, we adopt macro-average Precision (P), Recall (R), and F1 for the two tasks, while on Mastodon dataset, we ignore the neutral sentiment label in DSC task and for DAR task we adopt the average of the F1 scores weighted by the prevalence of each dialogue act.

\begin{table*}[t]
\centering
\fontsize{9}{12}\selectfont
\setlength{\tabcolsep}{1.6mm}{
\begin{tabular}{c|ccc|ccc|ccc|ccc}
\toprule
\multirow{3}{*}{Models} & \multicolumn{6}{c|}{Mastodon}                  & \multicolumn{6}{c}{DailyDialog}   \\ \cline{2-13}
            & \multicolumn{3}{c|}{DSC} & \multicolumn{3}{c|}{DAR} & \multicolumn{3}{c|}{DSC} & \multicolumn{3}{c}{DAR}\\\cline{2-13}
                        & P(\%)     & R(\%) & F1(\%) & P(\%) & R(\%)  & F1(\%)& P(\%)  & R(\%)  & F1(\%)& P(\%) & R(\%) & F1(\%)\\ \midrule
    JointDAS            &36.1   & 41.6  & 37.6  & 55.6  & 51.9  & 53.2  & 35.4  & 28.8  & 31.2  & 76.2  & 74.5  & 75.1  \\ \hline
     IIIM    & 38.7  & 40.1  & 39.4  & 56.3  & 52.2  & 54.3  & 38.9  & 28.5  & 33.0  & 76.5  & 74.9  & 75.7  \\  \hline
  DCR-Net& 43.2  & 47.3  & 45.1  & 60.3  & 56.9  & 58.6  & 56.0  & 40.1  & 45.4  & 79.1  & 79.0  & 79.1  \\  \hline
  BCDCN &38.2 &62.0 &45.9 &57.3 &61.7 &59.4 &55.2 &45.7 &48.6 &80.0 &80.6 &80.3\\ \hline
   Co-GAT & 44.0  & 53.2  & 48.1  & 60.4  & 60.6  & 60.5  & 65.9  & 45.3  & 51.0  & 81.0  & 78.1  & 79.4  \\ \hline
 \multirow{2}{*}{Co-GAT$^*$}& 45.40 & 48.11 & 46.47 & 62.55 & 58.66 & 60.54 & 58.04 & 44.65 & 48.82 & 79.14 & 79.71 & 79.39 \\
& $\pm$2.31 & $\pm$2.91 & $\pm$0.37 & $\pm$0.46 & $\pm$1.71 & $\pm$1.10 & $\pm$0.84 & $\pm$0.36 & $\pm$0.22 & $\pm$0.40&$\pm$0.16 &$\pm$0.14 \\ \midrule
 \multirow{2}{*}{DARER} & \textbf{56.04}$^\dag$ & \textbf{63.33}$^\dag$ & \textbf{59.59}$^\dag$ & \textbf{65.08}$^\ddag$ & \textbf{61.88}$^\dag$ & \textbf{63.43}$^\dag$ & \textbf{59.96}$^\ddag$ & \textbf{49.51}$^\dag$ & \textbf{53.42}$^\dag$ & \textbf{81.39}$^\dag$ & \textbf{80.80}$^\ddag$ & \textbf{81.06}$^\dag$ \\ 
 & $\pm$0.85 & $\pm$0.30 & $\pm$0.70 & $\pm$1.25 & $\pm$0.37 & $\pm$0.85 & $\pm$1.25 & $\pm$1.33 & $\pm$0.18 & $\pm$0.55&$\pm$0.43 &$\pm$0.04 \\ 
 \bottomrule
\end{tabular}}
\caption{Experiment results. $^*$ denotes we reproduce the results using official code. $\pm$ denotes standard deviation. \\$^\dag$ denotes that our DARER significantly outperforms Co-GAT with $p<0.01$ under t-test and $^\ddag$ denotes $p<0.05$.}
\label{table: results}
\end{table*}

\subsection{Implement Details and Baselines}
DARER is trained with Adam optimizer with the learning rate of $1e^{-3}$ and the batch size is 16.
We exploit 300-dimensional Glove vectors for the word embeddings, and the dimension of hidden states (label embeddings) is 128 for Mastodon and 256 for DailyDialog. 
The step number $T$ for recurrent dual-task reasoning is set to 3 for Mastodon and 1 for DailyDialog.
The coefficient $\gamma$ is set to 3 for Mastodon and $1e^{-4}$ for DailyDialog.
To alleviate overfitting, we adopt dropout, and the ratio is 0.2 for Mastodon and 0.3 for DailyDialog.
For all experiments, we pick the model performing best on validation set then report the average results on test set based on three runs with different random seeds.
The epoch number is 100 for Mastodon and 50 for DailyDialog.
All computations are conducted on an NVIDIA RTX 6000 GPU.

We compare our model with: JointDAS \cite{mastodon}, IIIM \cite{kimkim}, DCR-Net (Co-Attention) \cite{dcrnet}, BCDCN \cite{bcdcn} and Co-GAT \cite{cogat}.

\subsection{Main Results} \label{sec: mainresult}

Table \ref{table: results} lists the experiment results on the test sets of the two datasets.
We can observe that:\\
1. Our DARER significantly outperforms all baselines, achieving new state-of-the-art (SOTA).
In particular, over Co-GAT, the existing SOTA, DARER achieves an absolute improvement of 13.1\% in F1 score on DSC task in Mastodon, a relative improvement of over 1/4.
The satisfying results of DARER come from 
(1) our framework integrates not only semantics-level interactions but also prediction-level interactions, thus captures explicit dependencies other than implicit dependencies; 
(2) our SATG represents the speaker-aware semantic states transitions, capturing the important basic semantics benefiting both tasks;
(3) our DRTG provides a rational platform on which more effective dual-task relational reasoning is conducted.
(4) the advanced architecture of DARER allows DSC and DAR to improve each other in the recurrent dual-task reasoning process gradually. \\
2. DARER shows more prominent superiority on DSC task than DAR task.
We surmise the probable reason is that generally, act label is more complicated to deduce than sentiment label in dual-task reasoning.
For instance, it is easy to infer $u_i$'s Negative label on DSC given $u_{i}$'s Agreement label on DAR and $u_{i-1}$'s Negative label on DSC.
Reversely, given the label information that $u_i$ and $u_{i-1}$ are both negative on DSC, it is hard to infer the act label of $u_i$ because there are several act labels possibly satisfying this case, e.g., Disagreement, Agreement, Statement.\\
3. DARER's improvements on DailyDialog are smaller than those on Mastodon.
We speculate this is caused by the extremely unbalanced sentiment class distribution on DailyDialog.
From Fig. \ref{fig: label_distrib} we can find that over 83\% utterances do not express sentiment, while the act labels are rich and varied.
This hinders DARER from learning valuable correlations between the two tasks.

\subsection{Ablation Study}
\begin{table}[t]
\centering
\fontsize{9}{12}\selectfont
\setlength{\tabcolsep}{1.6mm}{
\begin{tabular}{c|cc|cc}
\toprule
\multirow{2}{*}{Variants} & \multicolumn{2}{c|}{Mastodon} & \multicolumn{2}{c}{DailyDialog} \\ \cline{2-5} 
                        & DSC            & DAR            & DSC              & DAR             \\ \midrule
        DARER         & \textbf{59.59}       &   \textbf{63.43}       &  \textbf{53.42}        &  \textbf{81.39}        \\ \hline
  w/o Label Embeddings   & 56.76        &   62.15       & 50.64         &  79.87              \\ 
  w/o Harness Loss                & 56.22         & 61.99      &   49.94        &   79.76        \\ \hline
     w/o SAT-RSGT        & 57.37        &  62.96      &  50.25        &  80.52        \\ 
     w/o DTR-RSGT       & 56.69       &   61.69       &  50.11          & 79.76          \\
 w/o TS-LSTMs       & 56.30      &   61.49      &   51.61         &  80.33 \\ \hline 
w/o Tpl Rels in SATG  & 58.23         &  62.21       &  50.99        &  80.70      \\ 
w/o Tpl Rels in DRTG  & 57.22       &   62.15      &   50.52        &  80.28     \\ 
  \bottomrule
\end{tabular}}
\caption{Results of ablation experiments on F1 score.} 
\label{table: ablation}
\end{table}
We conduct ablation experiments to study each component of DARER. Table \ref{table: ablation} lists the results.\\
(1)
Removing \textbf{label embeddings} causes prediction-level interactions cannot be achieved.
The sharp drops of results prove that our method of leveraging label information to achieve prediction-level interactions effectively improves dual-task reasoning via capturing explicit dependencies.
(2) Without \textbf{harness loss}, the two logic rules can hardly be met, so there is no constrain forcing DSC and DAR to gradually prompt each other, resulting in the dramatic decline of performances. 
(3) As the core of Dialog Understanding, \textbf{SAT-RSGT} captures speaker-aware semantic states transitions, which provides essential basic task-free knowledge for both tasks.
Without it, some essential indicative semantics would be lost, then the results decrease.
(4) The worst results of `w/o DTR-RSGT' prove that \textbf{DTR-RSGT} is the core of DARER, and it plays the vital role of conducting dual-task relational reasoning over the semantics and label information.
(5) The significant results decrease of `w/o TS-LSTMs' prove that \textbf{TS-LSTMs} also plays an important role in DARER by generating task-specific hidden states for both tasks and have some capability of sequence label-aware reasoning.
(6) Removing of the \textbf{temporal relations} (Tpl Rels) in SATG or DRTG causes distinct results decline.
This can prove the necessity and effectiveness of introducing temporal relations into dialog understanding and dual-task reasoning.
\subsection{Impact of Step Number $T$}
\begin{figure}[t]
 \centering
 \includegraphics[width = 0.46\textwidth]{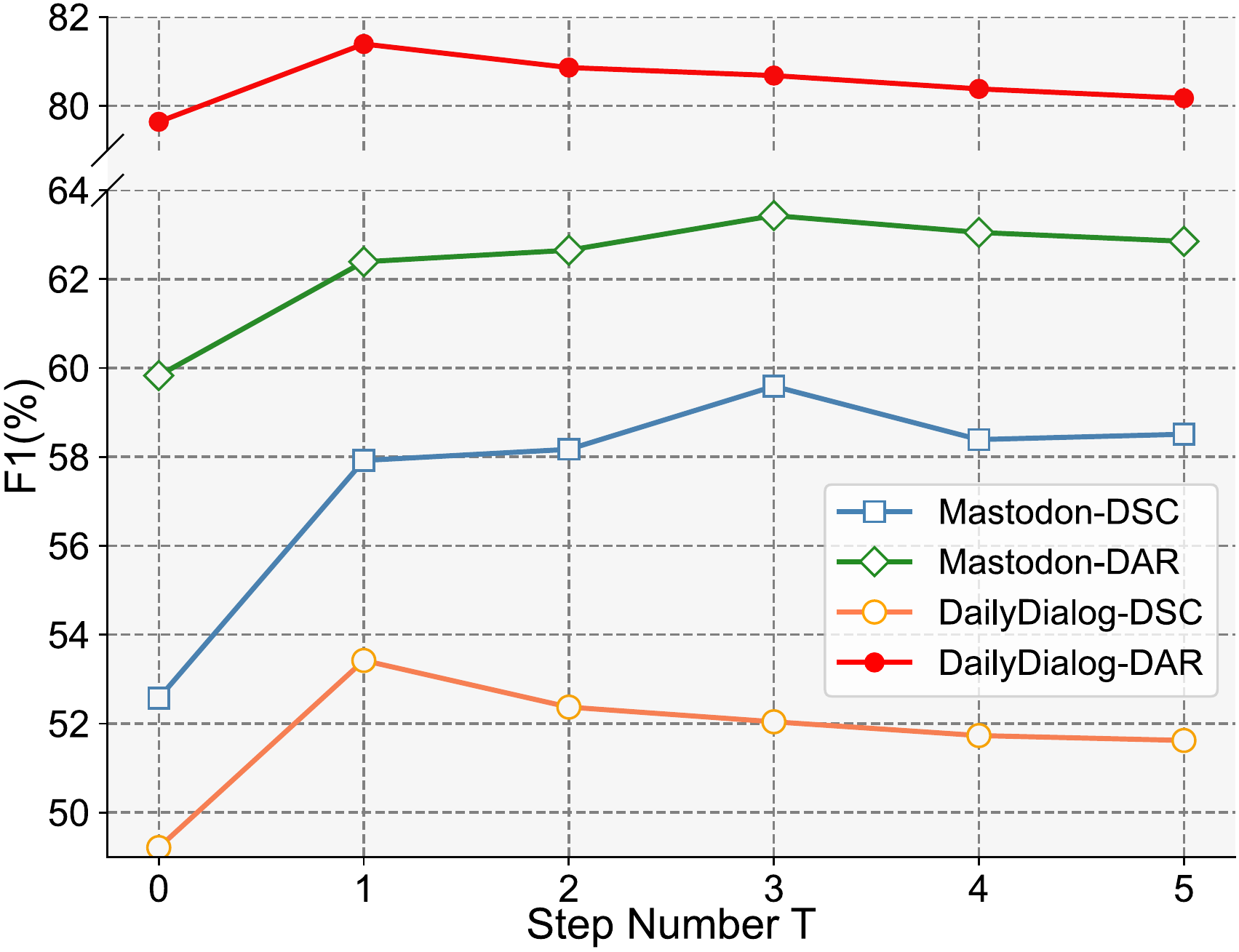}
 \caption{Performances of DARER over different $T$.}
 \label{fig: T}
\end{figure}

The performances of DARER over different $T$ are plotted in Fig. \ref{fig: T}.
$T=0$ denotes the output of Initial Estimation module is regarded as final predictions.
We can find that appropriately increasing $T$ brings results improvements.
Particularly, with $T$ increasing from 0 to 1, the results increase sharply.
This verifies that the Initial Estimation module can provide useful label information for dual-task reasoning.
Furthermore, DARER can learn beneficial mutual knowledge from recurrent dual-task reasoning in which DSC and DAR prompt each other.
Generally, when $T$ surpasses a certain point, the performances declines slightly.
The possible reason is that after the peak, more dual-task interactions cause too much deep information fusion of the two tasks, leading to the loss of some important task-specific features and overfitting.
\subsection{Effect of Pre-trained Language Model} 
\label{sec: ptlm}
\begin{table}[bh]
\centering
\fontsize{8}{11}\selectfont
\setlength{\tabcolsep}{1.3mm}{
\begin{tabular}{c|l|ccc|ccc}
\toprule
\multicolumn{2}{c|}{\multirow{3}{*}{\quad Models}} & \multicolumn{6}{c}{Mastodon}    \\ \cline{3-8}
         \multicolumn{2}{c|}{}   & \multicolumn{3}{c|}{DSC} & \multicolumn{3}{c}{DAR}\\ \cline{3-8}
                   \multicolumn{2}{c|}{}     & P(\%)     & R(\%) & F1(\%)    & P(\%)     & R(\%) & F1(\%)   \\ \midrule
\multirow{3}{*}{\rotatebox{90}{BERT}}
& + Linear         & 61.79 & 61.09 & 60.60 & 70.20 & 67.49 & 68.82 \\
& + Co-GAT           & 66.03 & 58.13 & 61.56 & 70.66 & 67.62 & 69.08 \\
& + DARER         & {65.98} & {67.39} & \textbf{66.42} & {73.82} & {71.67} & \textbf{72.73} \\ \midrule
\multirow{3}{*}{\rotatebox{90}{RoBERTa}}
&+ Linear     & 57.83 & 60.54 & 57.83 & 62.49 & 61.93 & 62.20 \\
& + Co-GAT        & 61.28 &57.25 & 58.26  &66.46 & 64.01 & 65.21 \\
& + DARER      & {61.36} & {67.27} & \textbf{63.66} & {70.87} & {68.68} & \textbf{69.75} \\ \midrule
\multirow{3}{*}{\rotatebox{90}{XLNet}}    
&+ Linear   & 61.42 & 67.80 & 63.35 & 67.31 & 63.04 & 65.09 \\
& + Co-GAT           & 64.01 & 65.30 & 63.71 & 67.19 & 64.09 & 65.60 \\
& + DARER         & {68.05} & {69.47} & \textbf{68.66} & {72.04} & {69.63} & \textbf{70.81} \\ \bottomrule
\end{tabular}}
\caption{Results based on different PTLM encoders.}
\label{table:ptlm}
\end{table}
In this section, we study the effects of three PTLM encoders: BERT \cite{bert}, RoBERTa \cite{roberta} and XLNet \cite{xlnet}, which replace the BiLSTM utterance encoder in DARER.
We adopt the base versions of the PTLMs implemented in PyTorch by \citet{transformers}.
In our experiments, the whole models are trained by AdamW optimizer with the learning rate of $1e^{-5}$ and the batch size is 16.
And the PTLMs are fine-tuned in the training process.
Results are listed in Table \ref{table:ptlm}.
We can find that since single PTLM encoders are powerful in language understanding, they obtain promising results even without any interactions between utterances or the two tasks.
Nevertheless, stacking DARER on PTLM encoders further obtains around 5\% absolute improvements on F1.
This is because our DARER achieves prediction-level interactions and integrates temporal relations, which complement the high-quality semantics grasped by PTLM encoders.
In contrast, Co-GAT only models the semantics-level interactions, whose advantages are diluted by PTLM.
Consequently, based on PTLM encoders, Co-GAT brings much less improvement than our DARER.

\subsection{Computation Efficiency}
\begin{table}[ht]
\centering
\fontsize{8}{12}\selectfont
\setlength{\tabcolsep}{1.6mm}{
\begin{tabular}{c|c|c|c|c}
\toprule
Models  & \begin{tabular}[c]{@{}l@{}}Number of \\ Parameters\end{tabular} & \begin{tabular}[c]{@{}l@{}}Training Time\\ per Epoch\end{tabular} & \begin{tabular}[c]{@{}l@{}}GPU \\ Memory\end{tabular}  & Avg. F1  \\ \midrule
Co-GAT  &6.93M        & 2.35s   &2007MB  &  53.66\% \\ \hline
DARER & 2.50M       & 2.20s  & 1167MB &  61.51\%  \\ \hline \hline
\textbf{Improve} & \textbf{-63.92}\%   &    \textbf{-6.38}\% &\textbf{-41.85}\%   & \textbf{14.63}\%  \\ \bottomrule
\end{tabular}}
\caption{Comparison with SOTA on different aspects.}
\label{table: computing_efficient}
\end{table}
In practical application, in addition to the performance, the number of parameters, the time cost, and GPU memory required are important factors.
Taking Mastodon as the testbed, we compare our DARER with the up-to-date SOTA (Co-GAT) on these factors, and results are shown in Table \ref{table: computing_efficient}. Avg. F1 denotes the average of the F1 scores on the two tasks.
Remarkably, although our DARER surpasses SOTA on Avg. F1 by 14.6\%, it cut the number of parameters and required GPU memory by about 1/2.
This is due to the parameter sharing mechanism in DARER.
Moreover, our DARER costs less time for training.
Therefore, it is proven that our DARER is more efficient in practical application.

\section{Related Works}\label{sec: relatedwork}

Dialog Sentiment Classification \cite{cmn,dialoggcn,ket,agru,topicdrivenerc,ercdag} and Dialog Act Recognition \cite{dar_2001,dac_selfatt,speakerawarecrf,Emotionaideddac} are both utterance-level classification tasks.
Recently, it has been found that these two tasks are correlative, and they can work together to indicate the speaker's more comprehensive intentions \cite{kimkim}.
With the development of well-annotated corpora, \cite{dailydialog,mastodon}, in which both the act label and sentiment label of each utterance are provided, several models have been proposed to tackle the joint dialog sentiment classification and act recognition task.

\citet{mastodon} propose a multi-task framework based on a shared encoder that implicitly models the dual-task correlations.
\citet{kimkim} integrate the identifications of dialog acts, predictors and sentiments into a unified model.
To explicitly model the mutual interactions between the two tasks, \citet{dcrnet} propose a stacked co-interactive relation layer and \citet{bcdcn} propose a context-aware dynamic convolution network to capture the crucial local context.
More recently, \citet{cogat} propose Co-GAT, which applies graph attentions on a fully-connected undirected graph consisting of two groups of nodes corresponding to the two tasks, respectively.

This work is different from previous works on three aspects.
First, we model the inner- and inter-speaker temporal dependencies for dialog understanding.
Second, we model the cross- and self-task temporal dependencies for dual-task reasoning;
Third, we achieve prediction-level interactions in which the estimated label distributions act as important and explicit clues other than semantics. 
\section{Conclusion and Future Work}\label{sec: conclusion}
In this paper, we present a new framework that integrates prediction-level interactions to leverage estimated label distribution as explicit and important clues other than implicit semantics.
Besides, we design the SATG and DRTG to introduce temporal relations into dialog understanding and dual-task reasoning.
Moreover, we propose a novel model named DARER to allow temporal information, label information, and semantics to work together to let DSC and DAR gradually promote each other, which is further forced by the proposed logic-heuristic training objective.
Experimental results demonstrate the superiority of our method, which not only surpasses previous models on performances by a large margin but also significantly economizes computation resources.

Our work brings two insights for dialog understanding and multi-task reasoning in dialog systems: (1) exploiting the temporal relations between utterances for reasoning; (2) leveraging estimated label distributions to capture explicit correlations;.
In the future, we will apply our method to other multi-task learning scenarios in dialog systems.

\section*{Acknowledgements}
This work was supported by Australian Research Council  Grant (DP180100106 and DP200101328).
Ivor W. Tsang was also supported by A$^*$STAR Centre for Frontier AI Research (CFAR).

\normalem
\bibliography{anthology}
\bibliographystyle{acl_natbib}




\end{document}